\documentclass[letterpaper, 10 pt, conference]{ieeeconf}  % Comment this line out if you need a4paper

\IEEEoverridecommandlockouts                              % This command is only needed if 
\usepackage{multirow}
\usepackage{tabularx}
\usepackage[font=small,labelfont=bf]{caption}
\usepackage{amsmath}
\usepackage{color}
\usepackage[normalem]{ulem}
\usepackage{hyperref}
\usepackage{nomencl}
\usepackage[ruled,vlined]{algorithm2e}
\usepackage{amsmath,amssymb,amsfonts}
\usepackage{algorithmic}
\usepackage{graphicx}
\usepackage{textcomp}
\usepackage{wrapfig}
\usepackage{booktabs}
\usepackage{mwe}
\usepackage{soul}
\usepackage[table, dvipsnames]{xcolor}
\usepackage{array,booktabs}

\usepackage{cite}
\usepackage{tikz,booktabs,xcolor}
\usepackage{cleveref}
\usepackage{listings}
\definecolor{light-gray}{gray}{0.95}
\usepackage{tcolorbox}
\usepackage[autostyle]{csquotes}
\usepackage{color}
\definecolor{darkblue}{rgb}{0.0,0.0,0.6}
\definecolor{maroon}{rgb}{0.5,0,0}
\definecolor{darkgreen}{HTML}{BAFFC9}
\definecolor{test_color3}{rgb}{0.0,0.0,0.0}
\definecolor{green}{rgb}{0,1,0}
\definecolor{test_color1}{HTML}{B5C0D0}
\definecolor{test_color2}{HTML}{8478BF}
\definecolor{red}{rgb}{1.0,0.1,0.0}
\definecolor{captionbgcolor}{rgb}{0.9, 0.9, 0.9} % gray

\DeclareCaptionFormat{colorbox}{%
  \begin{tcolorbox}[
    colback=light-gray,
    on line, 
    boxsep=1pt, % Padding between frame and content
    left=0pt, % Padding left
    right=0pt, % Padding right
    top=0pt, % Padding top
    bottom=0pt, % Padding bottom
    boxrule=0.5pt, % Border line width
    arc=3pt, % Corner radius
    auto outer arc, % Automatically use the arc size for straight corners
    nobeforeafter, % Avoid extra space before and after the tcolorbox
    width=\columnwidth,
  ]
  \centering #3
  \end{tcolorbox}%
}

\lstdefinelanguage{XML}
{
  basicstyle=\ttfamily,
  morestring=[s]{"}{"},
  morecomment=[s]{!--}{--},
  commentstyle=\color{test_color3},
  moredelim=[s][\color{black}]{>}{<},
  stringstyle=\color{red},
  identifierstyle=\color{red},
  belowcaptionskip=2pt,
  abovecaptionskip=-7pt,
}
\DeclareUnicodeCharacter{221A}{\ensuremath{-}}
\title{\LARGE \bf
SiSCo: Signal Synthesis for Effective Human-Robot Communication Via Large Language Models
}

\author{Shubham Sonawani, Fabian Weigend and Heni Ben Amor% <-this % stops a space
\thanks{S.~Sonawani, F.~Weigend, and H.~Ben~Amor are with the School of Computing and Augmented Intelligence, Arizona State University
{\tt\small \{sdsonawa, fweigend, hbenamor\}@asu.edu}}%
}

\begin{document}

\maketitle
\thispagestyle{empty}
\pagestyle{empty}

%%%%%%%%%%%%%%%%%%%%%%%%%%%%%%%%%%%%%%%%%%%%%%%%%%%%%%%%%%%%%%%%%%%%%%%%%%%%%%%%
\begin{abstract}
% For signal generation, best LLM is all you need!
Effective human-robot collaboration hinges on robust communication channels, with visual signaling playing a pivotal role due to its intuitive appeal. Yet, the creation of visually intuitive cues often demands extensive resources and specialized knowledge. The emergence of Large Language Models (LLMs) offers promising avenues for enhancing human-robot interactions and revolutionizing the way we generate context-aware visual cues. To this end, we introduce \textbf{SiSCo}--a novel framework that combines the computational power of LLMs with mixed-reality technologies to streamline the creation of visual cues for human-robot collaboration. Our results show that SiSCo improves the efficiency of communication in human-robot teaming tasks, reducing task completion time by approximately 73\% and increasing task success rates by 18\% compared to baseline natural language signals. Additionally, SiSCo reduces cognitive load for participants by 46\%, as measured by the NASA-TLX subscale, and receives above-average user ratings for on-the-fly signals generated for unseen objects. To encourage further development and broader community engagement, we provide full access to SiSCo's implementation and related materials on our GitHub repository.\footnote{\url{https://github.com/ir-lab/SiSCo}}
%Through a series of empirical investigations focusing on both human-robot collaboration and human-computer interaction, we evaluate the efficacy of SiSCo.
 %This study delves into the application of LLMs, particularly those proficient in generating Scalable Vector Graphics (SVG), for producing relevant visual signals within mixed-reality settings. 
% Through a series of empirical investigations focusing on human-robot collaboration and human-computer interaction, we evaluate the efficacy of SiSCo. Our results indicate that, leveraging LLM for the generation of visual signals, SiSCo significantly elevates the task efficiency of communicatory exchanges in human-robot teaming by approximately 60\% beyond that afforded by baseline natural language signals. Alongside this boost in signal efficacy, SiSCo contributes to a marked reduction in cognitive load for human subject, substantiated by a significantly lower NASA-TLX subscale scores for visual signals compared to baseline and, as whole system, robust score of 82 on the System Usability Scale. For broader community engagement and further innovation, we provide comprehensive access to SiSCo's implementation and related materials in our dedicated GitHub repository: \href{https://github.com/ir-lab/SiSCo}{https://github.com/ir-lab/SiSCo}.
\end{abstract}

\section{Introduction}

% \begin{enumerate}
%     \item Benefits and usefulness of LLMs
%     \item Mixed-Reality framework for signaling techniques to improve human robot collaboration
%     \item Visual signaling vs other signaling technique
%     \item LLM for context understanding and visual signal synthesis and its application in human robot interaction
% \end{enumerate}

% Signals and their types, Which signals are effective and their pro-cons
% In human robot collaboration, seamless and intuitive interaction requires intelligent and clear mode of communication that helps human-partner understand the robot intentions clearly. Different modes of communication signals such as text instructions, auditory, gestures and visual \cite{bonarini2020communication} have been investigated for their communication efficacy during human robot interaction. It has been shown that visual signals are the most effective, interpretable and least confusing mode of signals for human robot collaborative task\cite{lemasurier2021methods} while other form of signals are less intuitive. Specifically, visual signals used in mixed-reality have shown instrumental benefits in improving human robot commmunication [cite]. these visual signals are generally vector graphics images that can be visualized in different form of virtual, augment and mixed realities. 
% , successful interaction critically
% , to facilitate clear and intuitive information exchange during their interactions with robots
Human-robot collaboration (HRC) relies on clear and intuitive \emph{communication} channels between the human participants and their robotic counterparts. This foundational clarity is essential for both entities to understand each other's intentions and, in turn, engage in safe and effective interactions. Building upon this insight, a diverse set of communication modalities has been explored in the field of HRC, including text, auditory cues, gestures, and visual signals~\cite{bonarini2020communication}. Among these modalities, visual signals stand out due to their ability to (a) quickly capture attention and (b) instantly transmit a multitude of information. Accordingly, in recent years several approaches have been proposed that leverage visual cues within a mixed-reality environment to enhance HRC~\cite{andersen2016projecting, choi2022integrated,  gonzalez2023mixed, rivera2023toward } 
% \cite{andersen2016projecting, choi2022integrated, gonzalez2023mixed, rivera2023toward}. 
The proliferation of affordable output devices for virtual and augmented reality (e.g., Meta Quest, Microsoft HoloLens, Apple Vision Pro) has further increased the interest in visual forms of communication for human-robot teaming. %Often represented as vector graphics, these visual cues are particularly versatile, finding applicability across a spectrum of environments including virtual, augmented, and mixed reality \cite{citationneeded}.

% However, generation of these visual, especially image based, signals are intricate and time consuming task for human expert. In human centered computing, there is complete field of research called \emph{Visual Languages} that investigate the efficient and most impactful way of visual information generation. This reveals that visual signals may be efficient and intuitive in communicating the intentions in collaborative task but lacks direct or easy generation. 

While visual signals excel in quickly conveying a variety of information, their design and production are neither straightforward nor effortless and usually require human experts. The field of human-centered computing encompasses a dedicated area of study known as \emph{Visual Languages}~\cite{520784}, which focuses on principles for creating impactful visual information. In general, the process of creating these visual cues requires careful consideration in order to balance the need for clarity and intuitiveness against the resources and expertise needed for their development. Previous approaches for visual signaling in robots resorted to a predefined grammar~\cite{ganesan2018better}, i.e., a context-free language, or special-purpose authoring tools for the production of visual cues~\cite{10.1145/3610977.3634972}. While intuitive, these approaches do not allow for the extemporaneous generation of novel visual signals.

% Particularly, the ability of LLMs to understand the context given prompt by human and provide coherent and intelligent text based responses have created hype among the general population [cite]. 

On the other hand, recent work on Large Language Models (LLMs) shows the impressive ability in communicating with a human in a free-form fashion, i.e., without the need for a strict grammar or template. Particularly notable is the capacity of LLMs to comprehend the nuances of context in prompts provided by users and to generate coherent, contextually relevant textual responses~\cite{raj2023analyzing,sok2023chatgpt} which showcases LLMs reasoning capabilities. LLMs draw their capabilities from training on expansive textual datasets, comprising a diverse array of internet sources, including web pages, code repositories, and scholarly publications. By assimilating this wide spectrum of human-expressed knowledge, the models effectively encapsulate vast informational breadth into their trained weights. 
% These LLMs are trained on large text data available on internet web-pages, code repositories and literature which allows encapsulation of human knowledge in the form neural-network. 

\begin{figure}
    \centering
    \includegraphics[width=\linewidth]{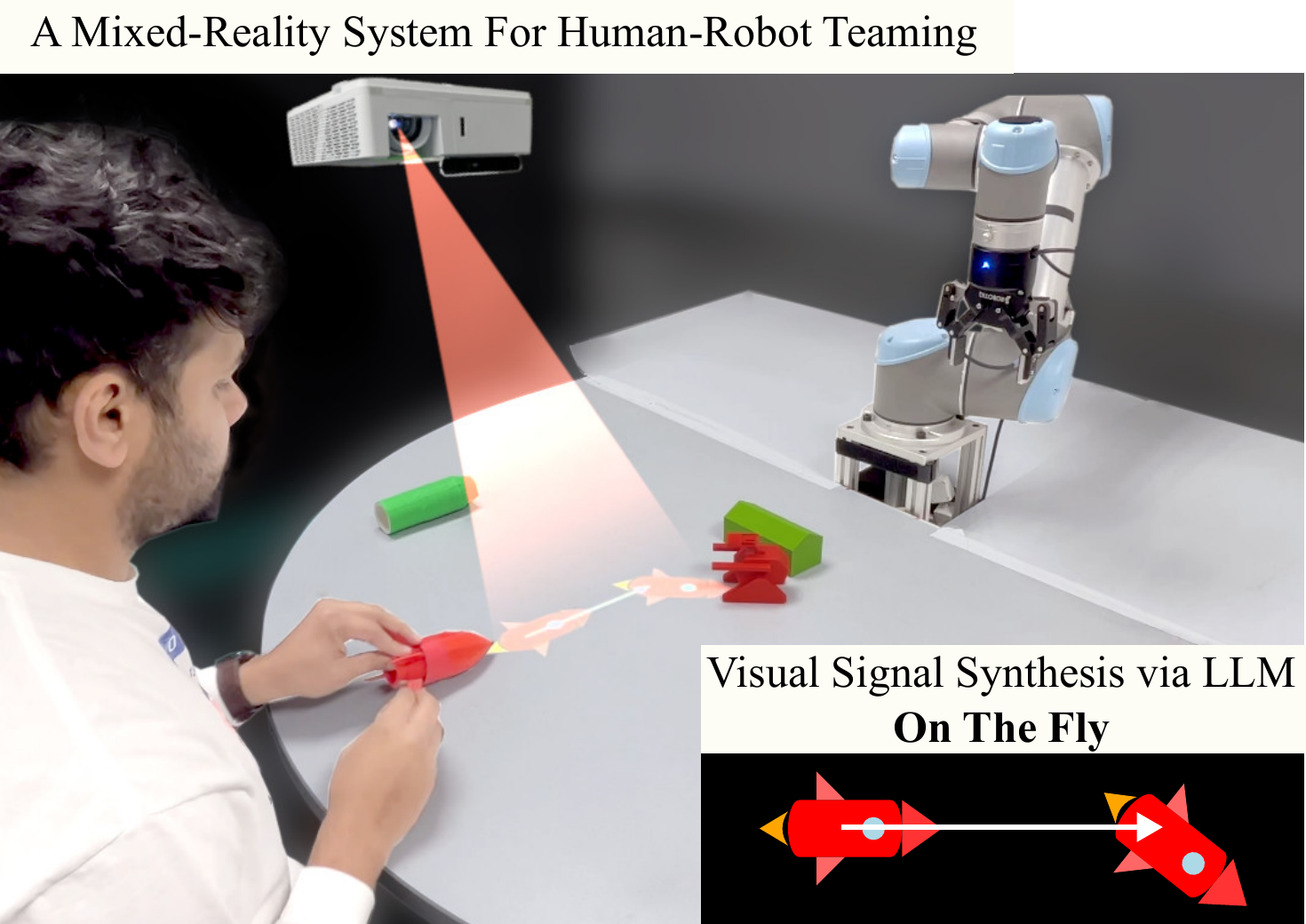}
    \caption{A participant engaging in a human-robot teaming task, with the SiSCo framework mediating by delivering visual signals via a mixed reality interface.}
    \label{fig:intprov2}
    \vspace{-15pt}
\end{figure}

\begin{figure*}[th]
    \centering
    \includegraphics[width=\linewidth]{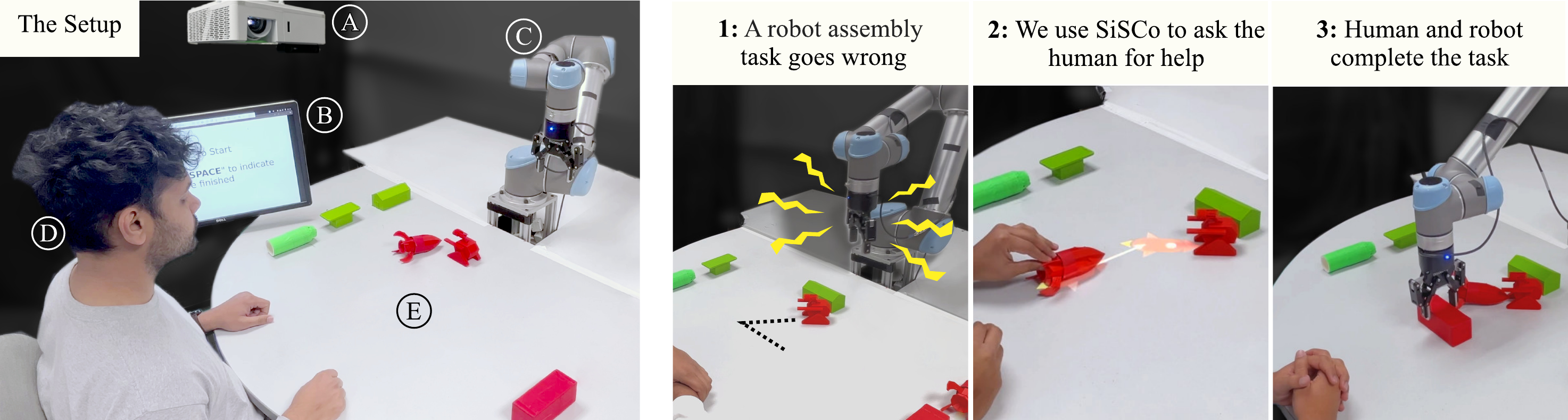}
    \caption{\textbf{Left:} The physical setup of the teaming task: The robot places objects on the tabletop surface environment. When the robot needs help, it uses SiSCo to present synthesized signals through a projector (A) or a monitor (B) to the human. \textbf{Right:} The task procedure during the human-robot teaming task.}
    \label{fig:setup}
\vspace{-10pt}
\end{figure*}

% Link of LLM to signals
In this paper, we address the question of whether it is possible to bridge the divide between textual and visual communication for the purposes of efficient human-robot teaming. Given a large language model, is there way we can leverage the embedded human knowledge to expertly generate signals for human robot collaboration? This paper specifically investigates the potential of LLMs to synthesize both natural language and visual signals that can adapt to changes in the environment, thereby enabling them to effectively convey robot intentions to users. An illustrative example of SiSCo deployed in a human-robot teaming scenario is presented in Figure \ref{fig:intprov2}.

To this point, the contributions made by this paper are outlined as follows:
\begin{itemize}
    \item The development of \textbf{SiSCo} (\textbf{Si}gnal \textbf{S}ynthesis for Effective Human-Robot \textbf{Co}mmunication), a novel framework integrating Large Language Models with mixed reality to produce legible visual signals on the fly, enhancing human-robot collaborative tasks.
    
    \item An empirical study involving human participants to rigorously assess the impacts of \textbf{SiSCo} on the enhancement of communication in teaming task performance, the alleviation of cognitive load, and the demonstration of its strong generalization capabilities in varied user inputs.
    
    \item Provision of the open-source repository for the \textbf{SiSCo} framework, including full system implementation details, to encourage widespread adoption and  allow for reproducibility.
   
\end{itemize}

\section{Related Work}

Effective communication plays a crucial role in human-robot interaction (HRI), with signaling methods serving as a critical component in clarifying intentions and reducing ambiguities~\cite{pascher2023communicate, bonarini2020communication}. Robots communicate with their human counterparts through a variety of signals, predominantly motion cues, gestures, and sounds. The clarity or \enquote{legibility}~\cite{dragan2013legibility} of these signals is essential to facilitate understanding of robot behavior and intentions.

Several studies have shown that both implicit and explicit signaling methods can significantly improve the communication gap in HRI tasks~\cite{dragan2013legibility, che2020efficient}. For example, implicit cues such as haptic feedback or eye movements can be particularly useful in close-proximity physical tasks~\cite{brosque2021collaborativewelding,cabrera2021cohaptics, neto2019gesture}. Implicit cues can be communicated alongside the original robot behavior but may require an adaptation phase, i.e., the human partner may have to learn the meaning of a given cue and how to best respond to it. By contrast, explicit cues provided via vision or language may require an additional communication step but are often easier to interpret for a human partner, e.g., the robot interrupting the task to provide a distress signal via language or through a picture~\cite{carissoli2023mental}. Immersive technologies, including virtual and mixed reality, have been adopted to deliver high-fidelity visual information and improve collaboration between humans and robots~\cite{williams2018virtual}. While virtual reality has the potential to increase the effectiveness and efficiency of collaborative tasks, it may cause user fatigue or nausea~\cite{chang2020virtual}. By contrast, mixed-reality techniques like projecting visual signals onto the real-world workspace can facilitate a simpler and more efficient communication mode in collaborative settings~\cite{andersen2016projecting}. All such approaches, however, require consistent and clear visual signals to achieve the intended effect~\cite{rokhsaritalemi2020review,olshannikova2015visualizing,dunston2005mixed}. The most common way to create visual signals for a new scenario is to design them  in a manual process. In turn, they can be reused as long as the task domain remains same. Alternatively, the work by \cite{ganesan2018better} proposed the concept of a domain-specific visual language, which utilizes a set of composable visual signals. Central to this proposition was the introduction of a robust grammar that served as an underlying generation mechanism. While the grammar itself stands as a well-conceived framework for consistent interpretation of signals, its application imposes an inherent rigidity. Specifically, visual signals, as defined by the established grammar, remain static and cannot be adjusted as the task and the environment evolves.

% Previous work \cite{ganesan2018better} proposed domain specific visual language as form of various type of visual signals where they defined grammar as logic to retain spatial information and affirmative actions. However, limitations comes to generalization ability as though grammar for visual signaling language is well established but visual signals stays constant and do not adapt for changes in objects and scenario.

This stands in stark contrast to the flexibility of LLMs which have recently led to a revolution in human-centered computing~\cite{wu2022ai, kim2024language}. LLMs provide a non-rigid communication interface, that allows humans to interact with a large corpus of textual data. Recent work, has shown that such textual interaction can also be leveraged to engage with robot partners~\cite{ahn2022can, huang2022inner}. Most importantly, LLMs have demonstrated an impressive adaptability to perform various tasks in a few-shot or even zero-shot manner. Accordingly, they are able to adapt to new tasks with minimal or no human effort. In the remainder of this paper, we investigate how the flexibility of LLMs can be used to generate visual signals for robotics tasks in a systematic and reliable manner.

\section{Methodology}
We present \textbf{Si}gnal \textbf{S}ynthesis for Effective Human-Robot \textbf{Co}mmunication (SiSCo) -- a system that utilizes LLMs to understand the task context and synthesize meaningful visual signals. Contextual information about the state of the collaboration task such as environmental descriptors (e.g., table dimensions) and problem statements (e.g., potential assembly errors by the robot) serve as input to the system. In turn, a visual representation is synthesized and projected into the scene to affect human behavior. While the projection can be performed via any display device (e.g., head-up display), we focus in the remainder of this paper on a projection-based methodology. \Cref{fig:setup} depicts the overall physical setup of our system.

\begin{figure}[t]
    \centering
    \includegraphics[width=\linewidth]{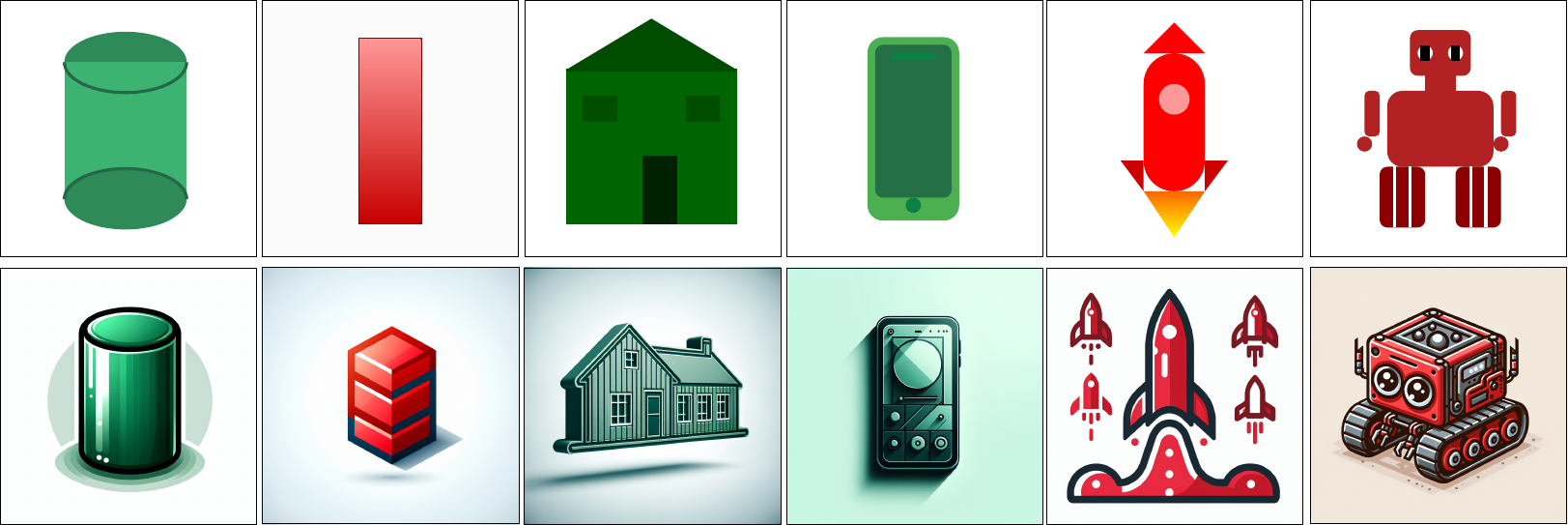}
    \caption{Object signals generated from SiSCo (Top Row) and Dalle-3 (Bottom Row) for same input prompt} %generated by Task Manager}
    \label{fig:sisco_vs_dalle}
\vspace{-10pt}
    
\end{figure}

Central to our approach is the generation of Scalable Vector Graphics (SVG) code, which is proficiently produced by the LLM framework for the creation of visual signals. The use of LLM is strategic, given its adeptness at contextual understanding and its ability to provide only the most pertinent information. While Vision Foundation Models~\cite{ramesh2022hierarchical} offer comprehensive visual representations of objects and scenes, they deliver an abundance of data in the form of high-dimensional RGB information, which may surpass the necessity for recognizing basic objects like a toy humanoid robot on a work surface (\Cref{fig:sisco_vs_dalle}). In stark contrast, SVG not only provides an efficient abstraction of these high-resolution images through simple geometric shapes and colors but also represents a human-interpretable file format. In addition, SVG files have the added advantage of being infinitely scalable without loss of quality; they can be rasterized to match the full dimensional range of traditional RGB arrays as required, ensuring compatibility with diverse display resolutions and optimizing system resource utilization for real-time applications in human-robot interaction.

%With the capability to generate signals reliably established, we proceeded to conduct a comparative analysis between SiSCo-generated signals and those from a Vision Foundation Model (VFM) known as DALL·E 3. The intent of this comparison was to examine the quality and efficiency of SiSCo's output in relation to a recognized industry standard. Presented in Figure~\ref{fig:sisco_vs_dalle}, the row-wise comparison underscores that SiSCo-generated signals not only exhibit a similarity in fidelity to VFM's output but also confer the added advantage of streamlining complexity. SiSCo's signal representation manages to encapsulate the essential visual information while avoiding the extensive dimensionality and superfluous detail inherent to VFM outputs. This reduction in unnecessary complexity could potentially lead to more efficient processing and lower computational costs without compromising the functional integrity of the encoded signals.

\subsection{System Overview}

% \subsection{Intention Projection System V2}
\label{subsec:intprov2}
% To this point, we leverage our prior work called intention projection \cite{andersen2016projecting,, which utilizes stereo setup of projector and camera to project signals on real-world canvas. LLM framework is used as mediator between robot and intention projection system. Motion planner communicates the relevant object properties information to VSS which help projector to overlay the real-world with synthesized visual signals. To render visual signals, we leverage combination of scalable vector graphics (SVG) and OpenCV library.  Robot operating system (Noetic) in tandem with Kinematics and Dynamic Library (KDL) is used to communicate between robot actions and projection visual signal.

%We designed our SiSCo framework such that LLM can be leveraged for synthesizing visual and natural signals. %These signals improve communication in HRC tasks. %  Our system comprises several key components: (1) 
A key component of our system is a mixed-reality setup, known as intention projection~\cite{sonawani2023projecting,sonawani2022xrrob,sonawani2022vam}, to communicate synthesized visual signals to the human. The setup employs a projector to superimpose visual information onto the real-world setting for clear communication of intentions and instructions. %(2) A UR5 robot, which serves as the collaborative agent within our teaming task. (3) 
\Cref{fig:setup} explains overall setup where the user, labeled as D, cooperates with the Universal Robot (UR5), indicated as C, to complete an assembly task on a table, designated as E. The objects required for the assembly are positioned on both sides of the table. If the robot experiences a malfunction, it initiates a query to the SiSCo framework. SiSCo then processes this query and conveys the synthesized visual signals to the user. These signals can be displayed using two methods: a projector (A) targeting the tabletop surface, enhancing spatial relevance, or an adjacent monitor (B) for a natural or visual signals.

The teaming task in this environment require cooperation between a robot, acting as the assembler, and a human, serving as an observer and assistant. In the example procedure depicted on the right in \Cref{fig:setup}, the goal is to arrange the objects in a Z-shaped structure. In \Cref{fig:setup} (Step-1), the robot has already placed two objects, but intentionally (simulated) failed to pick up the rocket. SiSCo then synthesizes a visual signal and projects it onto the table to ask the human for help (Step~2). The human places the rocket, which enables the robot to complete the Z-shaped structure (Step~3).

\subsection{Definitions}
\label{subsec:definitions}

\begin{figure}
    \centering
    \includegraphics[width=\linewidth]{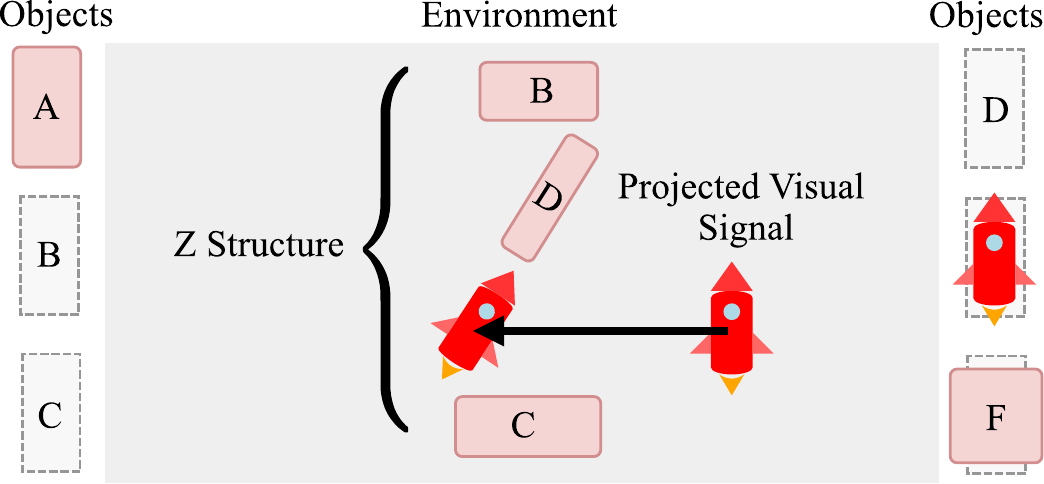}
    \caption{The teaming task as a schematic: The robot assembles a Z-shaped structure on the tabletop using the objects B, C, D and the red rocket. The robot placed the first three objects but malfunctioned when placing the red rocket. SiSCo projects a visual signal to instruct the human to assist.}
    \label{fig:problem}
\vspace{-10pt}
\end{figure}

% This section formalizes the teaming task and the workspace environment. Using the example in \Cref{fig:problem}, our definitions are as follows:
This section aims to formalize the components related to the teaming task, including the workspace, interacting objects, and structures. Below we provide definitions for each of these components:

\textbf{Environment:} 
% We define the environment as a rectangular 1.4\,m$\times$0.7\,m workspace on the tabletop surface. To provide mixed reality based visual signals we subdivided this realworld area into canvas of a $1400\times700$ grid (gray area in \Cref{fig:problem}). The robot assembles structures around the center of the environment using objects. 
We define the operational environment as a 1.4m by 0.7m rectangular workspace on a tabletop, further subdivided into a 1400 $\times$ 700 cell grid for visual signals provided by projection based Mixed Reality (MR), as illustrated in \Cref{fig:problem}. This grid, serving as a digital canvas, enables precise MR-based visual cues for robot and human interaction within this space. At the center of this digitally augmented environment, the robot is tasked with assembling structures from various objects, simulating complex interaction scenarios. This setup optimizes the use of space for human-robot collaboration and tests the robot's ability to interpret and act upon MR signals in real-time.

\textbf{Objects:} We position the objects at the left and right edges of the environment. Each physical object has a maximum width of 5 cm and a length of 13 cm. Their height varies between 6 cm and 9 cm. Additionally, each object has been assigned a description, denoted by $\Delta$, and a color, represented by $\Theta$.

\textbf{Structure:} One by one the robot picks and places objects to assemble alphabet-inspired structures in the center of the environment. In the example in \Cref{fig:problem}, the structure has a Z-shape.
%The tasks involve the 

\textbf{Problem Formalization:} While assembling the structure, the robot simulates having a problem and being unable to place an object at its intended goal position. We formalize the problem via six parameters. All parameter values are strings. The following example values define the Z-Problem in \Cref{fig:problem}, where the robot fails to place the rocket:

\begin{itemize}
    \item[$\Psi$]: structure (e.g. ``Z'')
    \item[$\Delta$]: object description (e.g. ``Rocket'')
    \item[$\Theta$]: object color (e.g. ``Red'')
    \item[$\Phi$]: goal position (e.g. ``[496, 100]'')
    \item[$\Pi$]: goal orientation (e.g. ``35 deg'')
    \item[$\Omega$]: instruction: (e.g. ``insert from right'')
\end{itemize}

From these parameters, SiSCo then synthesizes a signal to make the human complete the task for the robot (Projected Visual Signal in \Cref{fig:problem}). As a result, signals are synthesized to guide the human on how to contribute to resolving a problem within the environment.
% \vspace{-10pt}

 %At random point in time and goes to predefined stuck position. Once robot reaches predefined position, SiSCo is triggered with relevant information about next object to be assembled, which then generates signals to inform the human about how to assist in completing the structure. 
% Therefore, each generated signal is synthesized to inform the human on how to help resolving a problem in the environment. 
%Building upon our previous work in intention projection \cite{andersen2016projecting,sonawani2023projecting,sonawani2022xrrob,sonawani2022vam}, we employ a stereo configuration comprising both a projector and a camera to cast signals onto a real-world surface. The LLM framework operates as an intermediary, facilitating communication between the robotic agent and human partner via the intention projection system.
%The motion planner is tasked with transmitting crucial object property information to the SiSCo, which then assists the projector in augmenting the physical environment with synthesized visual cues. The visual signals are rendered utilizing a fusion of SVG and the OpenCV library for precision and visual clarity. To ensure seamless integration and communication between the robotic actions and the visual signal projection, we utilize the Robot Operating System (Noetic) in conjunction with the Kinematics and Dynamics Library (KDL). This integrated approach allows for an effective translation of robot actions into corresponding visual signals for real-time projection.

% ARCHITECTURE
%\subsection{Signals From Hierarchical Queries}
\subsection{Signal Synthesis}
\label{subsec:sfhq}

% SIGNAL TYPES
% To ask the human for help during human-robot teaming tasks, 
To communicate task goals and robot intentions to the human, 
SiSCo utilizes three signal modalities: Natural Language Signals (NLS), Visual Signals on a Monitor (VSM), or Visual Signals via Intention Projection (VSIntPro). 

\begin{figure}[t]
    \centering
    \includegraphics[width=\linewidth]{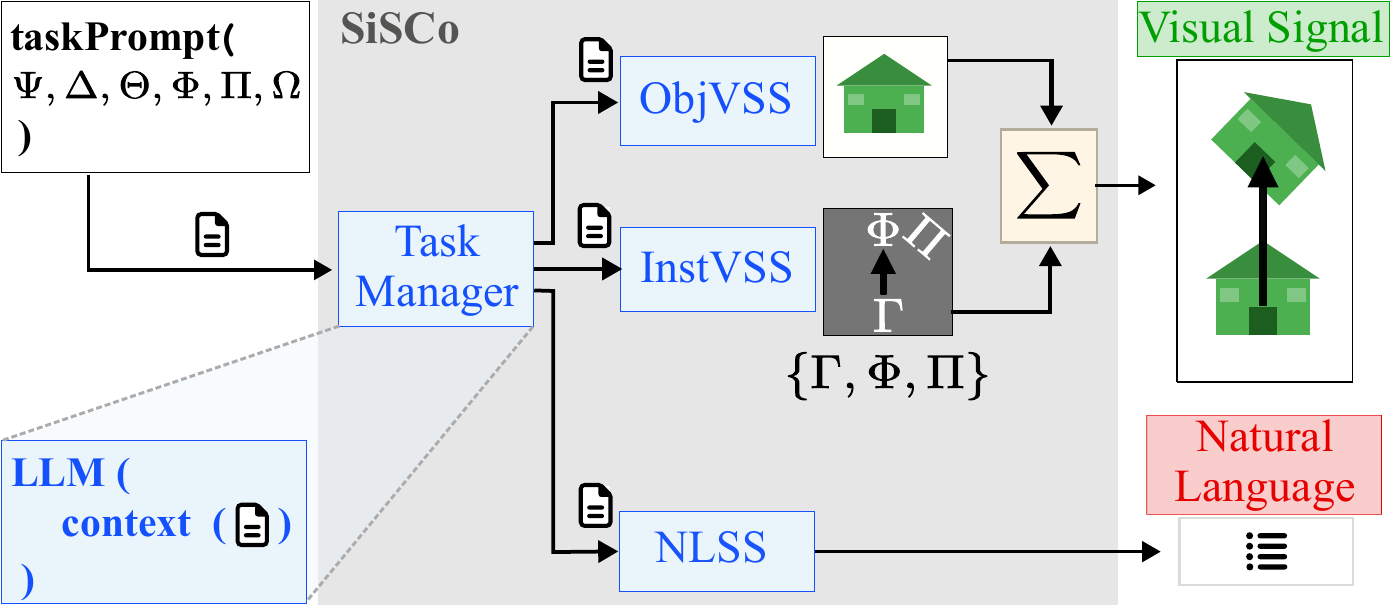}
    \caption{The architecture of our Signal Synthesizing Communication System (SiSCo). It takes in a task prompt and produces visual and natural language signals.}
    \label{fig:system}
    \vspace{-10pt}
    
\end{figure}

\begin{itemize}
    \item \textbf{Natural Language Signal (NLS)}: %Serving as a baseline for comparative analysis, 
    This mode leverages the LLM to generate textual instructions. The human participant is provided with a set of succinct on-screen directives that details task properties and the object to be manipulated.
    % as delineated in the Taskmanager section in \ref{subsec:sfhq}. 
    \item \textbf{Visual Signal on Monitor (VSM)}: In this mode, SiSCo synthesizes a visual depiction of the required human intervention and displays it on the monitor. %objects already assembled, alongside the subsequent object that requires human handling. The monitor displays these representations to guide the participant's interactions. 
    
    \item \textbf{Visual Signal via Intention Projection (VSIntPro)}: This mode employs the mixed-reality setup to signify the forthcoming object and its intended direction of manipulation. The signals are projected directly onto the tabletop. 
\end{itemize}

The SiSCo system synthesizes these signals through hierarchical LLM queries and post-processing. As depicted in \Cref{fig:system}, we formalize the hierarchical processing pipeline as four LLM function calls. Namely, the function calls are: 1) A Task Manager to subdivide the incoming prompt for following function calls, 2) A Natural Language Signal Synthesizer (NLSS) to summarize the task prompt in four bullet points, 3) An Object Visual Signal Synthesizer (ObjVSS) to process a prompt into an object icon and 4) An Instruction Visual Signal Synthesizer (InstVSS) to process a prompt into a visual instruction.
 %from an input task prompt comprised of the problem parameters ($\Psi, \Delta, \Theta, \Phi, \Pi, \Omega$). %The formal definition of its inputs follow in \Cref{subsec:definitions}. 
% Internally, SiSCo synthesizes signals through hierarchical LLM queries and post-processing. 
All four LLM functions wrap a context around their input prompt, i.e., they prepend and append contextual information to the prompt before they query the model. Formally, we denote the prefix as $\mathrm{PRE}$ and the postfix as $\mathrm{POST}$. The entire prompts including prefix and postfix are provided in our github repository. We describe their functions in the following sections:
% \footnote{\url{https://github.com/ir-lab/SiSCo/core.py}}.
% Sloppy descriptions. I am also not sure if we should move the task definitions before the architecture sectionhttps://www.overleaf.com/project/65381ad33c25d7e1cbebb6ba

\textbf{Task Manager:} As depicted in \Cref{fig:system}, the input task prompt first reaches the Task Manager function, which then generates three distinct prompts for subsequent LLM function calls. The Task Manager wraps the pretext $\mathrm{PRE}_\mathrm{TM}$ and posttext $\mathrm{POST}_\mathrm{TM}$ around the input prompt and queries the LLM. Then, it uses regular expressions to parse the LLM return into three new prompts to be passed to NLSS, ObjVSS, and InstVSS.

\textbf{NLSS:} The Natural Language Signal Synthesizer (NLSS) takes the refined task prompt provided by the Task Manager and envelops it with $\mathrm{PRE}_\mathrm{NLSS}$ and $\mathrm{POST}_\mathrm{NLSS}$. The resulting query compels the LLM to summarize the task description into a succinct set of four bullet points to be presented to the human. 

% The Natural Language Signal Synthesizer receives the processed task prompt from the Task Manager, wraps $\mathrm{PRE}_\mathrm{NLSS}$ and $\mathrm{POST}_\mathrm{NLSS}$ around it, and queries the LLM. $\mathrm{PRE}_\mathrm{NLSS}$ and $\mathrm{POST}_\mathrm{NLSS}$ instruct the LLM to summarize the incoming prompt into four bullet points to be read by a human.

\begin{figure}[t]  
% \begin{minipage}{0.5\textwidth}
   %caption={Synthesized SVG code}]
\centering
%\vspace{-10pt}
\begin{tcolorbox}[width=\linewidth,left=0pt,right=0pt,top=0pt,bottom=0pt,boxrule=0.5pt]
    \begin{tiny}{
    \selectfont \fontsize{6.1}{4}
% \vspace{-5pt}
\begin{lstlisting}[language=XML] 
<svg width="250" height="250" xmlns="http://www.w3.org/2000/svg">
  <!-- White background -->
  <rect width="100%" height="100%" fill="white" />
  <!-- Robot arm base -->
  <rect x="100" y="200" width="50" height="20" fill="#c0c0c0" />
  <!-- Robot arm first segment -->
  <rect x="120" y="130" width="10" height="70" fill="#4682b4" />
  <!-- Robot arm second segment -->
  <rect x="120" y="80" width="10" height="50" fill="#4682b4" />
  <!-- Robot arm joint circles -->
  <circle cx="125" cy="130" r="5" fill="#c0c0c0" />
  <circle cx="125" cy="80" r="5" fill="#c0c0c0" />
  <!-- Robotiq two-finger gripper -->
  <rect x="115" y="60" width="20" height="20" fill="#c0c0c0" />
  <rect x="110" y="40" width="5" height="20" fill="#4682b4" />
  <rect x="135" y="40" width="5" height="20" fill="#4682b4" />
  <!-- Gripper fingers -->
  <rect x="110" y="30" width="5" height="10" fill="#c0c0c0" />
  <rect x="135" y="30" width="5" height="10" fill="#c0c0c0" />
  <!-- Gripper finger tips -->
  <circle cx="112.5" cy="30" r="2.5" fill="#c0c0c0" />
  <circle cx="137.5" cy="30" r="2.5" fill="#c0c0c0" />
</svg>
    \end{lstlisting}
    }
\end{tiny}
\end{tcolorbox}
% \end{minipage}
\caption{Raw SVG code output generated by the ObjVSS function.} %We shortened the code for visualization purposes, indicated by [...].}
\label{fig:svg_code}
\vspace{-10pt}

\end{figure}

% REST OF SVG

\begin{figure}[t]
    \centering
    % \vspace{5pt}
    \includegraphics[width=\linewidth]{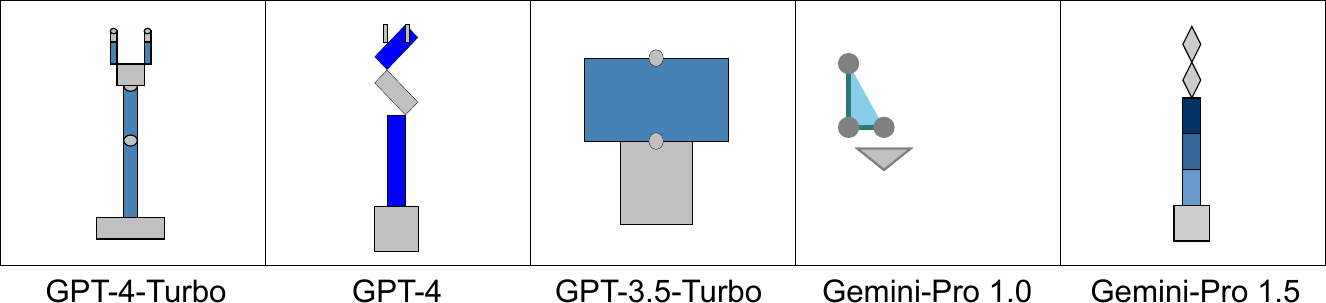}
    \caption{Comparison of visual signals generated by different LLMs. ObjVSS input prompt: \enquote{\textit{Generate an icon for an object with the description robotic arm with three degrees of freedom and parallel-jaw gripper and of blue and silver color.}}}
    \label{fig:llm_comparisons}
    % \vspace{-5pt
\end{figure}

\textbf{ObjVSS:} The Object Visual Signal Synthesizer (ObjVSS) generates the visual representation of the object in the form of SVG code. It receives a descriptive input prompt detailing the object description~($\Delta$) and color~($\Theta$) from the TaskManager. Subsequently, it augments the prompt with the context prefix $\mathrm{PRE}_\mathrm{ObjVSS}$ and postfix $\mathrm{POST}_\mathrm{ObjVSS}$, which delineate the specifications for the SVG creation, such as setting its target dimensions (e.g., 210$\times$210 pixels) and specifying the background color. Examples of ObjVSS outputs in the form of SVG code snippet is shown in \Cref{fig:svg_code} and are visualized in \Cref{fig:llm_comparisons}.
% Visual representation of object in the form of SVG code comes to reality in ObjVSS by taking clear descriptive prompt as input from Task Manager. The $\mathrm{PRE}_\mathrm{ObjVSS}$ and $\mathrm{POST}_\mathrm{ObjVSS}$ specifies parameters like dimensions e.g. 210$\times$210 and background color.

    % <svg width="250" height="250" xmlns="http://www.w3.org/2000/svg">
    %   <!-- Robotic arm base -->
    %   <rect x="100" y="200" width="50" height="50" fill="silver" />
    %   <!-- First degree of freedom -->
    %   <rect x="115" y="100" width="20" height="100" fill="blue" />
    %   <!-- Second degree of freedom -->
    %   <rect x="115" y="50" width="20" height="50" 
    %   transform="rotate(-45 25 75)" fill="silver" />
    %   <!-- Third degree of freedom -->
    %   <rect x="115" y="0" width="20" height="50" 
    %   transform="rotate(45 125 25)" fill="blue" />
    %   <!-- Parallel-jaw gripper -->
    %   <rect x="110" y="0" width="5" height="20" 
    %   fill="silver" />
    %   <rect x="135" y="0" width="5" height="20" 
    %   fill="silver" />
    % </svg>
    
\textbf{InstVSS:} For the Instruction Visual Signal Synthesizer~(InstVSS), the Task Manager provides a prompt that encapsulates both the goal position ($\Phi$) and goal rotation ($\Pi$) of the object to be manipulated. The contextual information $\mathrm{PRE}_\mathrm{InstVSS}$ and $\mathrm{POST}_\mathrm{InstVSS}$ obligate the LLM to contrive 1) suitable start position ($\Gamma$) and goal position in pixel coordinates 2) the SVG code that visualizes the trajectory, and 3) the object's orientation expressed in degrees, indicating clockwise rotation from the vertical axis.    

\textbf{Sigma:} The amalgamation of visual signals is conducted by the Sigma ($\Sigma$) function. Specifically, on an empty black canvas, the SVG output from the ObjectVSS is superimposed onto the designated start ($\Phi$) and end goal ($\Gamma$) positions, as determined by the InstVSS. At the goal position, the SVG representation of the object is rotated by the specified angle $\Pi$. Subsequently, the trajectory SVG, also provided by the InstVSS, is superimposed to yield a comprehensive visual signal of the object's movement from the start to the goal position. 

% From initial tests whether SiSCo framework is feasible or not, an empirical analysis was conducted to determine which LLMs performs optimally for the generation of signals. This analysis is crucial to establish a sound rationale for selecting a particular LLM suited for the SiSCo framework. Here, various well-established LLMs were evaluated for their ability to understand the context and produce visual signals that are representative of input prompts. As illustrated in Figure~\ref{fig:llm_comparisons}, the generated svg image data visually indicates that OpenAI's GPT-4-Turbo outperforms its counterparts and Gemini from Google.
% To evaluate the viability of the SiSCo framework, a thorough empirical analysis was carried out with the specific aim to ascertain the most effective Large Language Models (LLMs) for generating visual signals. 
% The purpose of this analysis was not merely to identify performance differentials among LLMs but to provide a well-founded justification for the selection of a particular model.% best suited to the rigorous demands of the SiSCo framework. 

\begin{figure}[t]
    \centering
    \includegraphics[width=\linewidth]{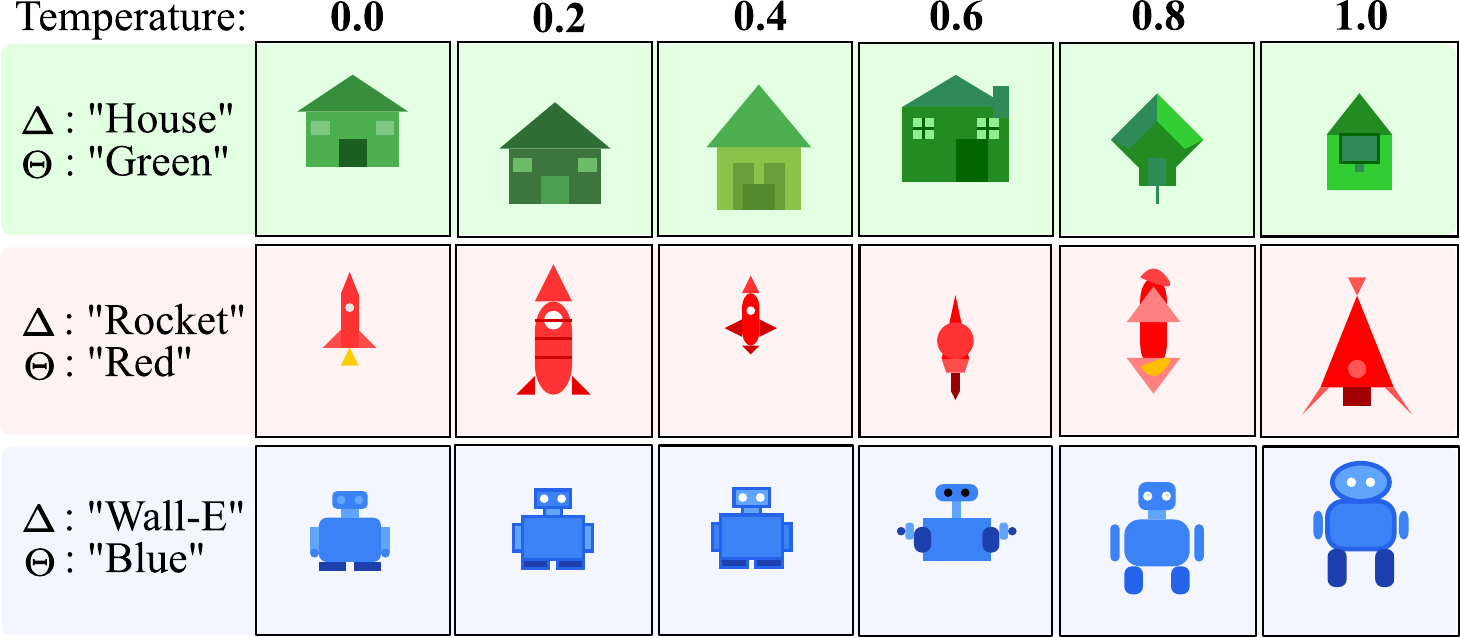}
    \caption{Object representations can be synthesized reliably even with high-temperature values.}
    \label{fig:temperature_object}
\vspace{-10pt}
\end{figure}

\subsection{LLM Selection}
% \vspace{-15pt}
Because SiSCo operates through text-based queries, it can interface with any LLM. We conducted initial experiments to select the most suitable and advanced LLM in the field for synthesizing reliable signals. \Cref{fig:llm_comparisons} depicts a representation of example outputs. It became evident from our initial comparison that, at the time of this work, OpenAI’s GPT-4-Turbo exceeded the signal generation capabilities of the other. This compelled us to select GPT-4-Turbo as the signal synthesis component within the SiSCo framework for this work.

%, facilitating a visual comparison of performance levels across the tested models. 
%Because all   %The following example demonstrates the synthesized SVG code generated by GPT-4-Turbo, showcasing its ability to comprehend the context and incrementally provide pertinent code comments and steps.
% Furthermore, for LLMs, \textit{temperature} \cite{chen2021evaluating, openaitemp} is parameter setting which allows the determinism in generated output, lower the \textit{temperature} less random the output will be and vise versa. Given the GPT-4 perform well at synthesis of signals, we also confirmed its generation reliability for different \textit{temperature} values. Specifically, Figure~\ref{fig:temperature} shows the \textbf{ObjVSS} ability to generate various object signals for given \textit{temperature} range. Once \textbf{ObjVSS} could generate reliable signals, we aslo verified the reliability of \textbf{InstVSS} together with \textbf{ObjVSS} for different \textit{temperature} values as shown in Figure~\ref{fig:temperature_object}. 
% \subsection{Increasing Temperature}
% Examples are depicted in \Cref{fig:temperature}

Acknowledging that GPT-4-Turbo demonstrates proficiency in the synthesis of signals, we further assessed its consistency in generating outputs across different \textit{temperature} settings. For LLMs, the temperature parameter influences the stochastic nature of the output, as detailed in \cite{chen2021evaluating, openaitemp}. A lower temperature setting results in more deterministic outputs, while a higher value facilitates greater randomness. 
The comparison depicted in \Cref{fig:temperature_object}  illustrates the ObjVSS outputs for three distinct inputs, each defined by object descriptions ($\Delta$) and color ($\Theta$), across a temperature range from 0.0 to 1.0. Even at the highest temperature setting, the generated objects remain clearly representative of the input descriptions. This outcome underscores the robustness of the carefully designed LLM ObjVSS, demonstrating its ability to maintain consistency and accuracy under varying conditions.

 %specifically showcases ObjVSS outputs for three different inputs of object description ($\Delta$) and color ($\Theta$) examples across a temperature range from 0.0 to 1.0. Even at the highest temperature setting, objects still recognizably represent the input object descriptions. This provide empirical evidence to the robustness of carefully engineered LLM ObjVSS.

\begin{figure}[t]
    % \vspace{5pt}
    \centering
    \includegraphics[width=\linewidth]{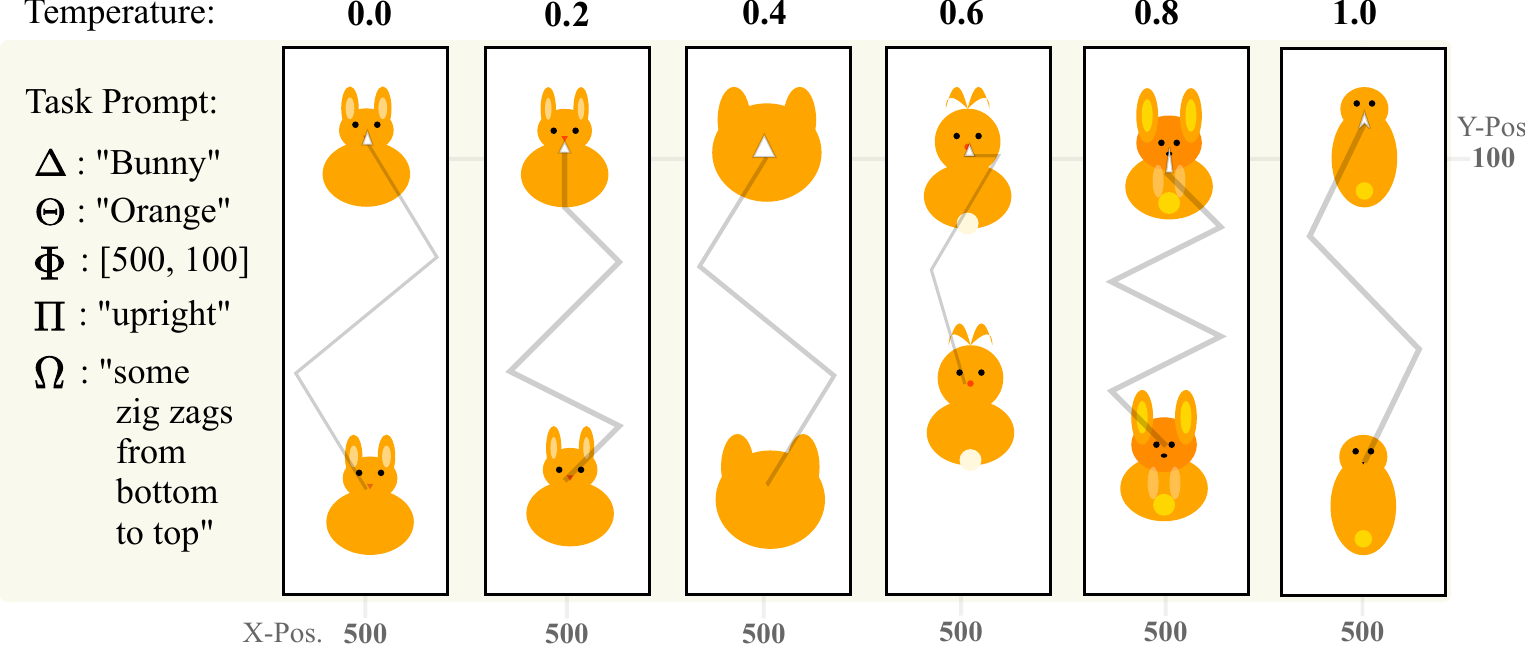}
    \caption{Visual signal synthesis stays robust with increasing temperature. The goal position at [500, 100] is matched, icons resemble a bunny, and the trajectory shows a zig-zag pattern.}
    \label{fig:temperature}
\vspace{-10pt}
\end{figure}

Furthermore, we illustrate the robustness of the ObjVSS and InstVSS functions by examining the merged visual signals in Figure \ref{fig:temperature}. Across various tested temperature settings, the target position ($\Phi$) at [500, 100] is consistently attained, icons effectively depict an orange ($\Theta$) bunny ($\Delta$), and the trajectory displays an upward zig-zag pattern ($\Omega$). Thus, we conclude that GPT-4-Turbo is well-suited for assessing the capabilities of signal synthesis for human-robot collaboration in our forthcoming experiments.
% Further, we showcase the robustness of the ObjVSS and InstVSS functions by comparing merged visual signals in \Cref{fig:temperature}. Across the range of tested temperature settings, the goal position ($\Phi$) at [500, 100] is matched, icons resemble an orange ($\Theta$) bunny ($\Delta$), and the trajectory shows an upward zig-zag pattern ($\Omega$). Therefore, we concluded that GPT-4-Turbo is suitable for evaluating the signal synthesis capabilities for human-robot collaboration in our following experiments.
% evaluated the \textbf{InstVSS} system in conjunction with \textbf{ObjVSS}. This assessment aimed to confirm the consistency of both components in generating reliable visual signals when subjected to various \textit{temperature} settings. The results of this comprehensive reliability check are illustrated in Figure~\ref{fig:temperature}.

\section{Experiments}
We designed an extensive human subject study to assess the quality of SiSCo for effective human robot communication. The study is subdivided into two parts: 1) A real-robot teaming task where a UR5 and a human assemble structures and 2) a visual signal evaluation questionnaire.
% \subsection{Human Subject Study}
\subsubsection{Human-Robot Teaming Task}

% The \Cref{tab:probem_definitions} summarizes all problems of our test set. We choose parameters such that they test the generalization capabilities of the LLM. For example, the orientation ``same'' equals to ``0 deg''.

As defined in \Cref{subsec:intprov2}, the teaming task involved assembling structures at the center of the tabletop environment. During assembly, the robot got stuck and used SiSCo to ask the human for help. For this task, we used a set of six distinct objects: They included a red rocket, a red cuboid, a red Wall-E robot, a green mobile phone, a green house, and a green cylinder. Using these objects, participants encountered six assembly problems defined in \Cref{tab:probem_definitions} in randomized order and with random SiSCo signal modalities (NLS, VSM, and VSIntPro). We randomized problems such that participants completed two problems for each signal modality. 

\begin{table}[h]
\centering
\footnotesize
\caption{Test set definitions for our practical teaming task}
\label{tab:probem_definitions}
\resizebox{\columnwidth}{!}{%
\begin{tabular}{cccccc}
\toprule
\multicolumn{1}{c}{\cellcolor[HTML]{EFEFEF} Struct.} \hspace{0.1em} & 
\multicolumn{2}{c}{\cellcolor[HTML]{EFEFEF} Object} \hspace{0.1em} & 
\multicolumn{3}{c}{\cellcolor[HTML]{EFEFEF} Placement \vspace{0.3em}}\\
\multicolumn{1}{c}{\cellcolor[HTML]{EFEFEF} $\Psi$} \hspace{0.1em} & 
\cellcolor[HTML]{EFEFEF} $\Delta$ (Desc.) & 
\multicolumn{1}{c}{\cellcolor[HTML]{EFEFEF} $\Theta$ (Col.) } \hspace{0.1em} & 
\cellcolor[HTML]{EFEFEF} $\Phi$ (Pos.) &  
\cellcolor[HTML]{EFEFEF} $\Pi$ (Ori.) & 
\cellcolor[HTML]{EFEFEF} $\Omega$ (Inst.) \vspace{0.3em} \\ 
S      & Cuboid       & Red      & 496, 262 & 90 deg     & from bottom          \\
Z      & Rocket       & Red      & 452, 306 & 45         & from bottom          \\
U      & Wall-E Robot & Red      & 396, 336 & same       & from left            \\
O      & Cylinder     & Green    & 598, 170 & no change  & from bottom          \\
R      & Mobile       & Green    & 612, 414 & -pi/4      & insert from right    \\
K      & House        & Green    & 496, 152 & 45 degrees & slide up from bottom \\ 

\bottomrule
\end{tabular}%
}
% \vspace{-5pt}
\end{table}

To evaluate the effect of SiSCo's signal modalities, we defined objective and subjective metrics. The objective metric one \textbf{OM1} captures failure and success rates, defining failure as: (1) manipulating the wrong object ($\Delta$ or $\Theta$), (2) placing the object with an incorrect orientation ($\Pi$), (3) failure to follow the placing instruction ($\Omega$), and (4) placement beyond 10\,cm from the goal position ($\Phi$). The second objective metric \textbf{OM2} quantifies task efficiency, measuring task completion time (duration from when the robot signaled being stuck to when the human placed the object, allowing the robot to continue) and comprehension time (duration from signal display to the participant touching an object). As the subjective metric \textbf{SM1} we utilized the NASA Task Load Index \cite{feick2020virtual} to assess the participant's perceived mental demand and frustration when following SiSCo generated signals. We asked participants to provided their NASA Task Load Index scores after they encountered a signal type (NLS, VSM, or VSIntPro) for the first time. Further, we facilitated the System Usability Scale (SUS) \cite{sus} as the subjective metric \textbf{SM2} to assess the perceived overall effectiveness of SiSCo in the human-robot teaming task. Participants provided their SUS scores after completing all six assembly tasks.

\subsubsection{Questionnaire}

\begin{figure}
    \centering
    \includegraphics[width=\linewidth]{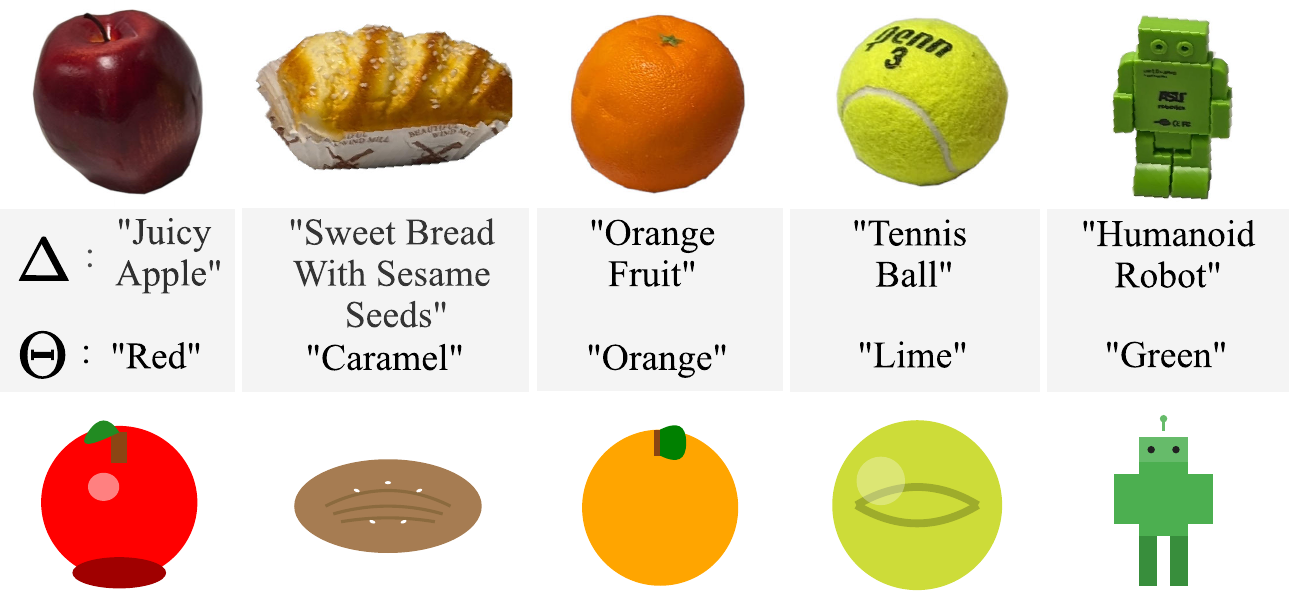}
    \caption{Real objects, prompt object descriptions ($\Delta$), prompt colors ($\Theta$), and generated visual signals. These objects were used for the second part of the questionnaire study}
    \label{fig:objects}
    \vspace{-10pt}
\end{figure}

As the second part of our experiment, we asked the human to complete a questionnaire, which further evaluated SiSCo's signal efficacy, quality, and user preference. Our questionnaire had four parts. The first question asked participants to rank the effectiveness of each interaction mode (NLS, VSM, and VSIntPro) by assigning a score from 1 for the most effective to 3 for the least effective. We used the ranking as our subjective metric \textbf{SM3}. 

The second questionnaire item assessed the object representation quality of ObjVSS outputs. We placed five real-world objects (depicted at the top in Figure~\ref{fig:objects}) in front of the user. Then, we presented generated ObjVSS representations (bottom of \Cref{fig:objects}) in random order. Humans had to select the real object that corresponded to the signal and rate the signal representation on a scale ranging from -5 (unrecognizable) to 5 (ideal representation). We utilized selection success rate as the objective metric \textbf{OM3} and the rating as the subjective metric \textbf{SM4}. %by asking participants to identify the intended object signal from several available options. Additionally, subjective metric \textbf{SM4} is obtained by inviting participants to rate the signal's quality  once they finish identifying the object.
The third section of the questionnaire asked participants to input a string that corresponds to an image combining both ObjVSS and InstVSS elements. Specifically, the image required participants to provide an input representing the action, \enquote{\textit{Insert a green-colored leek object from the left to the center of the image}.} This section was designed to evaluate whether participants could accurately generate a signal prompt based solely on the visual representation, effectively reversing the typical process. For metric \textbf{SM5}, participants were then instructed to rate the re-engineered signal derived from their interpretation of the image.
% The third section of the questionnaire prompted participants to enter string input values corresponding to an image that combines both ObjVSS and InstVSS elements. The image tasked participants with providing an input that represents the action, \enquote{\textit{Insert a green-colored leek object from the left to the center of the image}.} This section aimed to assess whether participants can accurately generate a signal prompt based solely on the visual representation. This metric helps whether SiSCo can be given to any human subject and generate reliable visual signals basically reversing the role. For metric \textbf{SM5}, participants were instructed to rate the re-engineered signal derived from their interpretation of the image. 

In the final part of the questionnaire, participants were granted complete control over SiSCo and were asked to input prompts for any generic object, orientation and trajectory of their choice. After receiving the generated signal from SiSCo, participants provided the subjective metric \textbf{SM6} by rating the accuracy with which the signal represents the specified properties. 

% \subsection{Hypotheses}

The underlying question behind the design of SiSCo is to evaluate its efficacy in improving communication between robot and human, and its ability to generate interpretable signals given novel inputs. To this point, we investigated the following hypotheses:
%We answer that by conducting subjective experiments and post-hoc analysis of the respective data obtained during experiment. 
\begin{itemize}
    \item [\textbf{H1}:] Visual signals synthesized from SiSCo improve task performance compared to natural language signals. % We define task performance metric inspired from \cite{de2022human, hopko2021effect} as combination of task accuracy: a task is performed accurately when the human recognizes all properties ($\Psi$, $\Delta$, $\Theta$, $\Phi$, $\Pi$, $\Omega$) from their visualized signals and task efficiency: efficiency is high when the human finishes the task faster, i.e., low task completion time and comprehension time. 
    % FW: Visual signals improve task performance compared to natural language signals. Similar to \cite{de2022human, hopko2021effect} we define task performance as a combination of task accuracy (success rate) and task efficiency (completion time). The success rate is high, when the human recognizes all task properties ($\Psi$, $\Delta$, $\Theta$, $\Phi$, $\Pi$, $\Omega$) correctly and completes the task.

    \item [\textbf{H2}:] The cognitive load is lower in visual signal based modes compared to the natural language signal mode.%  For cognitive load, we utilize metrics from NASA-TLX, specifically mental demand, temporal demand and frustration. In addition, ranking of different modes are obtained from subjects which also serves how effective the different mode was for the human. 
    
    \item [\textbf{H3}:] SiSCo generates %synthesizes
 human-interpretable representations for \textit{any} unseen input.
\end{itemize}

To provide evidence for or against the above hypotheses, we evaluate the objective and subjective metrics from the human-robot teaming tasks and questionnaires. The utilized metrics and their implications for the hypotheses are detailed in the following.

\section{Results and Analysis}

Our detailed experiments involved 21 participants, aged 18 to 36, including four females and seventeen males. The study received approval from the Institutional Review Board (IRB) under the ID STUDY00019583. We defined three independent variables for SiSCo's signal types: NLS, VSM, and VSIntPro.%—to investigate their effects on SiSCo's communication performance in accordance with the proposed hypotheses.

\subsection*{\textbf{H1:} Task Performance}
To evaluate task performance, we utilized \textbf{OM1} for the measurement of task accuracy and \textbf{OM2} to assess task efficiency, considering both as dependent variables. For analyzing differences in task success rates across different signal modalities, we treated the data from \textbf{OM1} as binary categorical measurements and applied Fisher's exact test to determine statistical significance. The results indicated a significant difference in success rates between VSM and NLI (\textit{stat} = 0.25, \textit{p} $<$ 0.05), and between VSIntPro and NLI (\textit{stat} = 0.17, \textit{p} $<$ 0.05). However, the difference between VSIntPro and VSM was not statistically significant (\textit{stat} = 0.7, \textit{p} $>$ 0.05). Despite this, data presented in \Cref{tab:obj_met} shows that VSIntPro outperformed NLI by 18\% and VSM by 3\% in the overall task success rate. To assess task efficiency, we analyzed comprehension time and task completion time based on \textbf{OM2} data to ascertain any significant differences. The distribution of the \textbf{OM2} data was found to be non-normal as confirmed by a normality test \cite{d1973tests}. Consequently, we employed the non-parametric Kruskal-Wallis test, a counterpart to the one-way ANOVA suitable for non-normal distributions, applying the Bonferroni correction to the \textit{p}-values to mitigate the risk of Type-I error.
% The results revealed significant differences in communication efficiency: for comprehension time, there was a noticeable disparity between VSIntPro and NLI (\textit{H} = 56.1, \textit{p} $<$ 0.05), between VSM and NLI (\textit{H} = 52.6, \textit{p} $<$ 0.05), and between VSIntPro and VSM (\textit{H} = 15.6, \textit{p} $<$ 0.05). While for the completion time trend followed similar to prior as VSIntPro and NLI (\textit{H} = 108.3, \textit{p} $<$ 0.05), between VSM and NLI (\textit{H} = 77.1, \textit{p} $<$ 0.05), and between VSIntPro and VSM (\textit{H} = 6.3, \textit{p} $<$ 0.05). Given this significant differences,  VSIntPro is the most effective communication mode among those evaluated.
The results revealed significant differences in communication efficiency. In terms of comprehension time, a clear disparity was observed between VSIntPro and NLI (\textit{H} = 56.1, \textit{p} $<$ 0.05), VSM and NLI (\textit{H} = 52.6, \textit{p} $<$ 0.05), and VSIntPro and VSM (\textit{H} = 15.6, \textit{p} $<$ 0.05). Similarly, the completion time exhibited a comparable trend, with significant differences between VSIntPro and NLI (\textit{H} = 108.3, \textit{p} $<$ 0.05), VSM and NLI (\textit{H} = 77.1, \textit{p} $<$ 0.05), and VSIntPro and VSM (\textit{H} = 6.3, \textit{p} $<$ 0.05). These findings indicate that VSIntPro is the most effective communication mode among those evaluated.
The aforementioned results show that SiSCo's visual signals significantly enhance task performance, thereby corroborating hypothesis \textbf{H1}. Furthermore, although VSIntPro did not significantly outperform VSM in task accuracy, participants ranked it as the superior mode of communication in the subjective questionnaire (\textbf{SM3} in Table~\ref{tab:sub_met}).
\begin{table}[t]
      \caption{Objective Metrics: Averaged across all participants}
        \centering
        \resizebox{\columnwidth}{!}{%
        \label{tab:obj_met}
        \begin{tabular}{rcccccc}        
        \toprule
        \rowcolor[HTML]{EFEFEF} 
        \multicolumn{7}{c}{\cellcolor[HTML]{EFEFEF} \textbf{OM1:} Teaming Task Success Rate (\%)\vspace{0.3em}} \\
        \rowcolor[HTML]{EFEFEF} &
        \multicolumn{1}{c}{$\Delta$ (Desc.)} &
        \multicolumn{1}{c}{$\Theta$ (Col.)} &
        \multicolumn{1}{c}{$\Phi$ (Pos.)} &
        \multicolumn{1}{c}{$\Omega$ (Inst.)} &
        \multicolumn{1}{c}{$\Pi$ (Ori.)} &
        \multicolumn{1}{c}{All \vspace{0.3em}} \\ 
        \cellcolor[HTML]{EFEFEF} NLS & 97.6 & 100.0 & 73.8 & 66.7 & 73.8 & 82.4 \\
        \cellcolor[HTML]{EFEFEF} VSM & 97.6 & 97.6 & 92.9 & 92.9 & 90.5 & 94.3 \\
        \cellcolor[HTML]{EFEFEF} VSIntPro & 97.6 & 97.6 & 100.0 & 100.0 & 90.5 & 97.1 \\
        \midrule
        \rowcolor[HTML]{EFEFEF} 
        \multicolumn{7}{c}{\textbf{OM2:} Teaming Task Temporal Analysis (s)\vspace{0.3em}} \\
        \rowcolor[HTML]{EFEFEF} & \multicolumn{3}{c}{Comprehension Time} & \multicolumn{2}{c}{Completion Time} & \vspace{0.3em} \\
        \cellcolor[HTML]{EFEFEF} NLS       & \multicolumn{3}{c}{28.9 $\pm$ 14.6} & \multicolumn{2}{c}{42.9 $\pm$ 19.8} \\
        \cellcolor[HTML]{EFEFEF} VSM      & \multicolumn{3}{c}{5.7 $\pm$ 2.8}   & \multicolumn{2}{c}{15.5 $\pm$ 5.5} \\
        \cellcolor[HTML]{EFEFEF} VSIntPro & \multicolumn{3}{c}{3.7 $\pm$ 1.3}   & \multicolumn{2}{c}{11.4 $\pm$ 3.0} \\ 
        \midrule
        \rowcolor[HTML]{EFEFEF} 
        \multicolumn{7}{c}{\textbf{OM3:} Questionnaire Object Recognition Success Rate (\%)\vspace{0.3em}}  \\
        \rowcolor[HTML]{EFEFEF} & Tennis Ball & Apple & Bread & Robot & Orange & All \vspace{0.3em} \\
        \cellcolor[HTML]{EFEFEF} VSIntPro & 100.0 & 68.4 & 100.0 & 94.7 & 87.5 & 90.1\\
        \bottomrule
        \end{tabular}%
        }        
\vspace{-10pt}
        
\end{table}

\subsection*{\textbf{H2:} Cognitive Load}
In the human-robot teaming task, participants remained stationary on table, which led to the decision to omit the physical demand and effort components of the NASA-TLX. Furthermore, the performance subscale from NASA-TLX was excluded as participants could not evaluate their successful task completion at the conclusion of each experiment as only expert knows whether human participant succeeded or not. Thus, we concentrated on mental demand, temporal demand, and frustration as our dependent variables to assess cognitive load, which are documented as \textbf{SM1} shown in \Cref{tab:sub_met}. Statistical verification of data normality was conducted based on D'Agostino's tests \cite{d1973tests}, paving the way for the Multivariate Analysis of Variance (MANOVA). The results delineated significant cognitive load variations when comparing Visual Signals with Intention Projection (VSIntPro) and the Natural Language Interface (NLI) (\textit{F}(3,32) = 4.5, \textit{p} $<$ 0.05), as well as between Visual Signals with a Monitor (VSM) and NLI (\textit{F}(3,32) = 3.0, \textit{p} $<$ 0.05). However, the difference between VSIntPro and VSM was not statistically significant (\textit{F}(3,32) = 0.9, \textit{p} $>$ 0.05). On average, VSIntPro achieved 46\% lower score on NASA-TLX subscales compared to NLI showing its prowess in easing the communication.  In terms of effectiveness in conveying information, ranked by \textbf{SM3} (see Table~\ref{tab:sub_met}), VSIntPro attained the highest rank score (1.1 $\pm$ 0.5), followed by VSM (2.0 $\pm$ 0.4) and NLI (2.9 $\pm$ 0.3), indicating a preference among participants for VSIntPro as the most effective mode. With VSIntPro demonstrating the least cognitive load and achieving the highest effectiveness ranking, we confirm our hypothesis \textbf{H2}.

\begin{table}
    %\caption*{Evaluation Metrics}
    %\begin{minipage}{.545\linewidth}
      \caption{Subjective Metrics: Averaged across all participants}
        \centering
        \resizebox{\columnwidth}{!}{%
        \label{tab:sub_met}
        
        \begin{tabular}{wc{1.0cm}wc{1.0cm}wc{1.0cm}wc{1.0cm}wc{1.0cm}wc{1.0cm}}
        \toprule
          \rowcolor[HTML]{EFEFEF}  \multicolumn{6}{c}{\textbf{SM1:} NASA-TLX (0 to 20) \vspace{0.3em}} \\
          \rowcolor[HTML]{EFEFEF}      & \multicolumn{2}{c}{Mental Demand} &  \multicolumn{2}{c}{Temporal Demand} & \multicolumn{1}{c}{Frustration} \vspace{0.3em}\\
          \cellcolor[HTML]{EFEFEF}NLS  & \multicolumn{2}{c}{7.4 $\pm$ 4.9} &  \multicolumn{2}{c}{4.3 $\pm$ 3.8} & \multicolumn{1}{c}{5.0 $\pm$4.6} \\
          \cellcolor[HTML]{EFEFEF}VSM  & \multicolumn{2}{c}{4.7 $\pm$ 4.0} &  \multicolumn{2}{c}{3.9 $\pm$ 3.1} & \multicolumn{1}{c}{3.2 $\pm$ 3.3} \\
          \cellcolor[HTML]{EFEFEF}VSIntPro  & \multicolumn{2}{c}{2.9 $\pm$ 2.4} &  \multicolumn{2}{c}{3.4 $\pm$ 3.1} & \multicolumn{1}{c}{2.7 $\pm$ 2.6} \\
          \midrule
          % \cellcolor[HTML]{EFEFEF}  \multicolumn{1}{c}{NLS}  & \multicolumn{2}{c}{7.2 $\pm$ 5.1} &  \multicolumn{2}{c}{{6.4 $\pm$ 4.4} & \multicolumn{1}{c}{4.8 $\pm$ 4.5} \\
          \rowcolor[HTML]{EFEFEF} \multicolumn{3}{c}{\textbf{SM2:} SUS (0 to 100)}  \hspace{0.3em} & \multicolumn{3}{c}{\textbf{SM3:} Signaling Modes Ranking  (3 to 1)} \vspace{0.3em} \\
          \rowcolor[HTML]{EFEFEF} 
          \multicolumn{3}{c}{SiSCo}  \hspace{0.3em} & 
          \multicolumn{1}{c}{VSIntPro} &  
          \multicolumn{1}{c}{VSM} & 
          \multicolumn{1}{c}{NLS \vspace{0.3em}} \\
          \multicolumn{3}{c}{82.0} &
          \multicolumn{1}{c}{1.1 $\pm$ 0.5} & 
          \multicolumn{1}{c}{2.0 $\pm$ 0.4} & 
          \multicolumn{1}{c}{2.9 $\pm$ 0.3} \\
          \midrule
          \rowcolor[HTML]{EFEFEF} 
          \multicolumn{6}{c}{\textbf{SM4:} Object Recognition Rating (-5 to 5)} \vspace{0.3em} \\
          \rowcolor[HTML]{EFEFEF} &
          Tennis Ball&
          Apple&
          Bread&
          Robot&
          \multicolumn{1}{c}{Orange \vspace{0.3em}}\\
          \multicolumn{1}{r}{\cellcolor[HTML]{EFEFEF}{VSIntPro}} & 
          3.3 $\pm$ 2.0& 
          3.2 $\pm$ 1.8& 
          1.2 $\pm$ 2.9& 
          4.3 $\pm$ 1.6& 
          1.3 $\pm$ 2.8\\
          \midrule
          \multicolumn{3}{c}{\cellcolor[HTML]{EFEFEF} 
          \textbf{SM5:} Signal Re-Engineering} \hspace{0.3em} & 
          \multicolumn{3}{c}{\cellcolor[HTML]{EFEFEF} 
          \textbf{SM6:} Creative Inputs}  \vspace{0.3em}\\
          \multicolumn{3}{c}{\cellcolor[HTML]{EFEFEF} 
          Rating (-5 to 5)} \hspace{0.3em} & 
          \multicolumn{3}{c}{\cellcolor[HTML]{EFEFEF} 
          Rating (-5 to 5) \vspace{0.3em}}\\
          \multicolumn{1}{r}{\cellcolor[HTML]{EFEFEF} VSIntPro }& \multicolumn{2}{c}{2.3 $\pm$ 2.1} & 
          \multicolumn{1}{r}{\cellcolor[HTML]{EFEFEF} VSIntPro }& \multicolumn{2}{c}{2.2 $\pm$ 3.0} \\
          % \bottomrule
          \midrule
          \end{tabular}       
        }
\vspace{-10pt}
\end{table}

\subsection*{\textbf{H3:} Generalization}
In our pursuit to enhance human-robot collaboration, assessing SiSCo's capacity for handling various human inputs was crucial. We assessed this adaptability by integrating subjective metrics (\textbf{SM4-6}) for users to evaluate the visual signals SiSCo produced in response to new inputs, alongside an objective metric (\textbf{OM3}) that quantified how accurately participants identified target objects.

Despite initial analyses, user ratings deviated from a normal distribution, failing to fulfill the criteria set by D'Agostino's tests \cite{d1973tests}. Consequently, we employed the non-parametric Wilcoxon signed-rank test to determine if user ratings were significantly more favorable compared to a baseline rating of \enquote{0}, which is hypothesized as a pivot for potential improvement in a range from 0 to 5.

The Wilcoxon test affirmed that user ratings were significantly above the baseline for all subjective metrics (\textbf{SM4}: \textit{stat} = 2976, \textit{p} $<$ 0.05; \textbf{SM5}: \textit{stat} = 181.5, \textit{p} $<$ 0.05; \textbf{SM6}: \textit{stat} = 147.5, \textit{p} $<$ 0.05), suggesting a decided preference for the SiSCo-generated visual signals. In conjunction with these findings, \textbf{OM3} revealed that participants successfully identified the correct objects in, on avarage, 90.1\% of cases, illustrating SiSCo’s reliability. Collectively, these results substantiate hypothesis \textbf{H3}, affirming both the user satisfaction with SiSCo and its effectiveness in a generalized application context.

Additionally, the System Usability Scale (SUS) was administered to gauge user perceptions of SiSCo's adaptability for human-robot collaboration. With a score of 82.0, SiSCo significantly surpassed the average SUS benchmark, reflecting a robust endorsement of the system's usability by the participants.

\section{Discussion and Limitations}
% This paper show ability of LLM, specifically GPT-4-Turbo, to incorporate knowledge obtained from big-corpora of internet data and synthesize relevant vector graphics that can be utilize for human robot interaction task and mixed-reality. 

We introduced SiSCo, a novel framework designed to enhance communication in human-robot collaboration by integrating mixed-reality environments with LLMs. SiSCo generates visual signals and cues on the fly to convey robotic intent and to influence human intervention. In this work, we utilized GPT4-Turbo for signal generation, which causes a notable dependence on OpenAI's server infrastructure. As LLM research advances in the future, we will focus on fine-tuning a model for local execution to mitigate this limitation. 

%Our evaluation revealed that SiSCo exhibits considerable adeptness in generating signals appropriate for facilitating human-robot collaboration tasks. Our findings demonstrate that measured task performance was significantly improved (60.8 \% ) when utilizing visual signals generated through intention projection (VSIntPro) compared to simple natural language signals. Further, the cognitive load experienced by users in the VSIntPro mode was significantly reduced—by 46\% as per NASA-TLX subscales. Most notably, the decrease in cognitive load is accompanied by an above-average System Usability Scale (SUS) score of 82 for the SiSCo system. Altogether, SiSCo's adept signal generation and broad user satisfaction indicate very strong potential for future work in incorporating LLMs and mixed reality in human-robot collaboration.

Our evaluation of SiSCo has revealed remarkable adeptness in generating signals tailored for facilitating human-robot collaboration tasks. Even when generating visual signals for previously unseen objects, users rated the object representation quality as above average in all subjective metrics. Objective metrics underscored these findings with a significant enhancement in measured task performance, 73\% faster task completion time and 18\% higher task success rate, when utilizing visual signals generated through intention projection (VSIntPro) compared to simple natural language signals. Additionally, users reported a substantial 46\% reduction in cognitive load with the VSIntPro mode, as measured by NASA-TLX subscales. 
%Notably, this decrease in cognitive load is complemented by an above-average System Usability Scale (SUS) score of 82 for the SiSCo system. 
Taken together, the adeptness of SiSCo in signal generation and the high level of user satisfaction elicit a strong potential for future endeavors integrating LLMs and mixed reality in human-robot collaboration.

\bibliographystyle{IEEEtran}
\bibliography{references}

\end{document}